\documentclass[10pt,twocolumn,letterpaper]{article}

\usepackage[pagenumbers]{iccv} 

\definecolor{iccvblue}{rgb}{0.21,0.49,0.74}
\usepackage[pagebackref,breaklinks,colorlinks,allcolors=iccvblue]{hyperref}

\def\paperID{2538} 
\def\confName{ICCV}
\def\confYear{2025}

\title{Group-wise Scaling and Orthogonal Decomposition \\for Domain-Invariant Feature Extraction in Face Anti-Spoofing}

\author{Seungjin Jung$^{1}$
\qquad
Kanghee Lee$^{1}$
\qquad
Yonghyun Jeong$^{2}$
\and
Haeun Noh$^{1}$
\qquad
Jungmin Lee$^{1}$
\qquad
Jongwon Choi$^{1}$\thanks{Corresponding author. Email: \href{choijw@cau.ac.kr}{choijw@cau.ac.kr}}
\and \\
$^{1}$Chung-Ang University\qquad$^{2}$Naver Cloud
}
\usepackage{multirow}
\usepackage{graphicx}
\usepackage{booktabs}
\usepackage{kotex}
\usepackage{colortbl}

\begin{document}
\maketitle
\AddToShipoutPicture*{%
     \AtTextUpperLeft{%
         \put(0,30){
           \begin{minipage}{\textwidth}
              \footnotesize
              Preprint version; final version will be available at \url{https://openaccess.thecvf.com}\\
              The IEEE / CVF International Conference on Computer Vision (2025)\\
              Published by: IEEE \& CVF
           \end{minipage}}%
     }%
}
\begin{abstract}
Domain Generalizable Face Anti-Spoofing (DGFAS) methods effectively capture domain-invariant features by aligning the directions (weights) of local decision boundaries across domains. However, the bias terms associated with these boundaries remain misaligned, leading to inconsistent classification thresholds and degraded performance on unseen target domains.
To address this issue, we propose a novel DGFAS framework that jointly aligns weights and biases through Feature Orthogonal Decomposition (FOD) and Group-wise Scaling Risk Minimization (GS-RM).
Specifically, GS-RM facilitates bias alignment by balancing group-wise losses across multiple domains. FOD employs the Gram-Schmidt orthogonalization process to decompose the feature space explicitly into domain-invariant and domain-specific subspaces. By enforcing orthogonality between domain-specific and domain-invariant features during training using domain labels, FOD ensures effective weight alignment across domains without negatively impacting bias alignment.
Additionally, we introduce Expected Calibration Error (ECE) as a novel evaluation metric for quantitatively assessing the effectiveness of our method in aligning bias terms across domains. Extensive experiments on benchmark datasets demonstrate that our approach achieves state-of-the-art performance, consistently improving accuracy, reducing bias misalignment, and enhancing generalization stability on unseen target domains.
Code:~\href{https://github.com/SeungjinJung/GD-FAS}{https://github.com/SeungjinJung/GD-FAS}
\end{abstract}
\begin{figure}[t!]
        \centering
        \includegraphics[width=\linewidth]{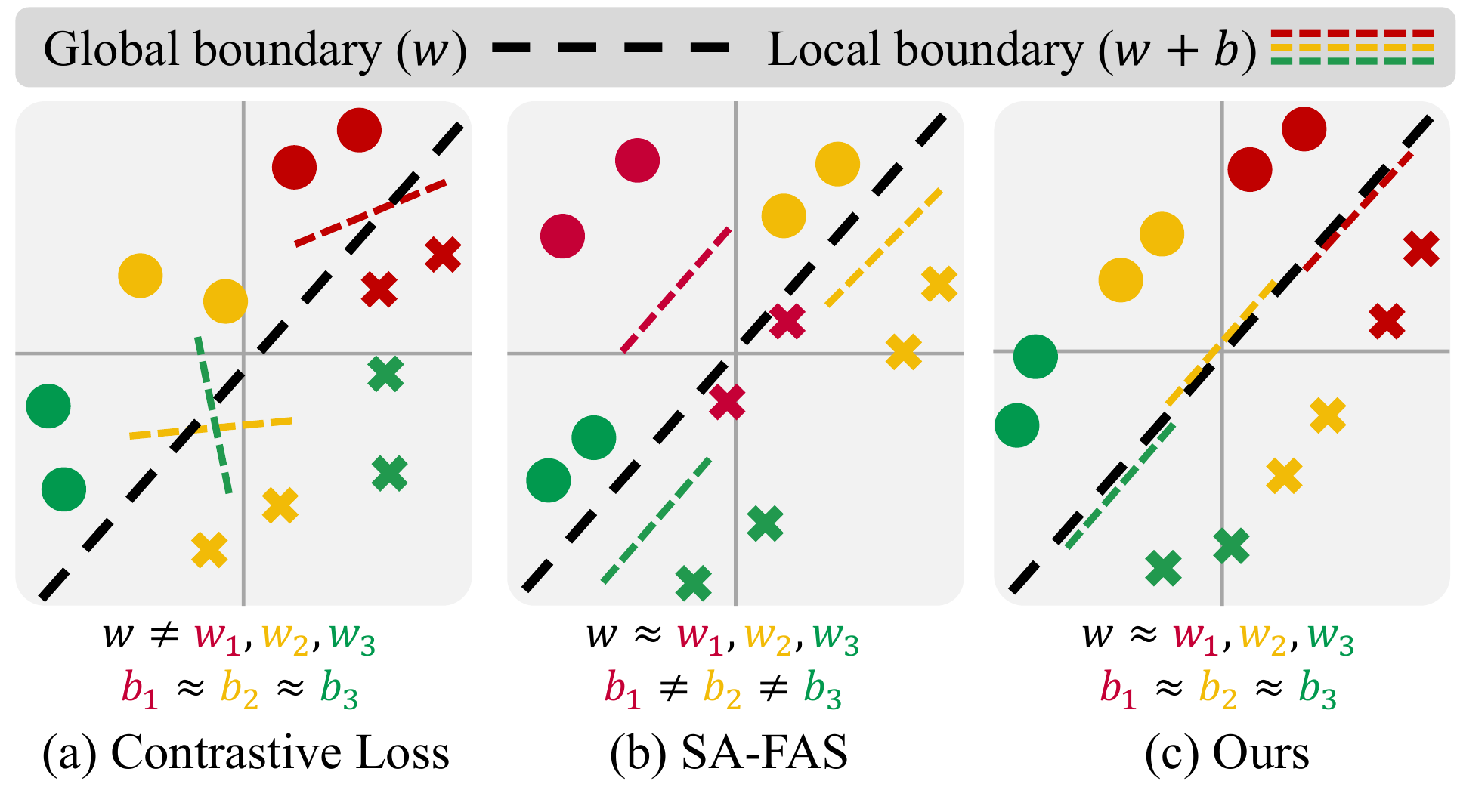}
\caption{{\bf{Bias term and weight term across domains}}:
Decision boundaries are defined by the classifier’s weights and biases. Domain invariance maintains consistent distinguishability across domains, typically by sharing a common weight term. In contrast, domain specificity represents domain gaps arising from variations in bias terms.
(a) Domain-wise contrastive loss enhances domain-specific separability with bias alignment but struggles to extract domain-invariant features.
(b) SAFAS aligns local decision boundaries for improved domain invariance, yet residual domain gaps persist, particularly due to bias misalignment.
(c) Our approach addresses this by jointly aligning weights and biases using Group-wise Scaling (GS) and Feature Orthogonal Decomposition (FOD), establishing a unified and consistent decision boundary.
}
\label{fig:teaser}
\end{figure}

\section{Introduction\label{sec:intro}}
With the rapid increase in the use of Face Recognition (FR)~\cite{Yang_2023_CVPR, Li_2023_CVPR} technology in critical applications such as payment systems and identity verification, the security of these systems has become more important than ever.
FR systems are widely used not only at airport entry points but also in personalized devices, making it critical for these systems to effectively respond to presentation attacks, where attackers employ technical methods to impersonate legitimate users.
In this context, Face Anti-Spoofing (FAS) technology has emerged as an essential defense mechanism to protect FR systems.

Recent FAS studies have focused on Domain Generalizable FAS (DGFAS) to overcome the cross-domain performance drop across various domains, especially the adaptability to new environments, camera sensors, and attack types unseen during training~\cite{liu2021dual,liu2021adaptive,wang2022domain,wang2022patchnet,sun2023rethinking, Srivatsan_2023_ICCV}.
Such DGFAS research primarily focuses on representation learning to extract domain-invariant features by training models to perform effectively with a single classifier.
One approach enhances representations by generating two distinct augmented patch images from the same image and aligning them in the feature space~\cite{wang2022patchnet}. Another approach segregates domain-specific features and aligns the directional consistency between spoofness and liveness~\cite{sun2023rethinking}. 
With the rise of Large Language Models (LLMs), recent studies have leveraged text embedding features extracted from LLMs as guidance to improve the extraction of domain-invariant features~\cite{Srivatsan_2023_ICCV, wang2025tf, liu2024cfpl, liubottom}.

Despite advancements in capturing domain-invariant features, DGFAS methods still encounter significant challenges arising from misaligned bias terms associated with local decision boundaries. As illustrated in Fig. 1-(a), contrastive loss enhances domain-specific separability and aligns bias terms across domains but fails to effectively extract domain-invariant features. In contrast, SAFAS~\cite{sun2023rethinking} aligns classifier weights across domains yet remains limited due to persistent bias misalignment, as depicted in \cref{fig:teaser}-(b). Addressing this challenge requires explicitly aligning the bias terms of local decision boundaries. A detailed discussion on resolving this bias misalignment issue is presented in \cref{sec:discussion}.

To address this challenge, we propose a novel DGFAS framework called GD-FAS, comprising Group-wise Scaling Risk Minimization (GS-RM) and Feature Orthogonal Decomposition (FOD).
GS-RM aligns the bias terms of classifiers by adaptively balancing group-wise losses across domains through a scaling factor applied at each training iteration. This approach ensures consistent separation of real and spoofed samples across multiple domains.
FOD decomposes feature representations into orthogonal domain-invariant and domain-specific subspaces using the Gram-Schmidt orthogonalization process. By constraining domain-specific features to their designated subspace, FOD aligns weight terms across domains without interfering with bias alignment, as shown in \cref{fig:teaser}-(c).
Consequently, our method effectively aligns bias and weight across different domains, substantially enhancing domain invariance and significantly improving cross-domain generalization performance.

Furthermore, we introduce Expected Calibration Error (ECE) as an evaluation metric to quantitatively assess the reliability and effectiveness of our method in aligning bias terms across different domains. By explicitly decomposing feature representations into orthogonal domain-invariant and domain-specific subspaces, our method achieves state-of-the-art performance, significantly enhancing accuracy, reducing bias misalignment, and improving generalization stability on unseen target domains.

The significant contributions of this research are as follows:

\begin{itemize} \label{contribution}
\item We propose a novel face anti-spoofing approach that ensures robust and consistent performance across diverse domains by decomposing into domain-invariant and domain-specific components orthogonally.
\item To enhance domain invariance, we introduce Group-wise Scaling Risk Minimization (GS-RM), which employs a loss-adaptive scaling factor to align classifier bias terms by balancing learning across domains. This ensures consistent and reliable classification of real and spoof samples in multiple domain scenarios.
\item To mitigate the domain gap, our approach introduces the Feature Orthogonal Decomposition (FOD) mechanism, which employs a contrastive loss guided by domain indices to construct domain-specific feature spaces. By enforcing orthogonality between domain-specific and domain-invariant features, this method effectively aligns the weight terms across distinct domains while leaving the bias term.
\item We introduce Expected Calibration Error (ECE) to quantitatively validate the effectiveness and reliability of bias term alignment across domains. Comprehensive experiments on standard benchmarks demonstrate that our method achieves state-of-the-art performance, consistently ensuring strong adaptability and stability in unseen target domains.
\end{itemize}
\section{Related Work}
\begin{figure*}[t]
        \centering
        \includegraphics[width=1\linewidth]{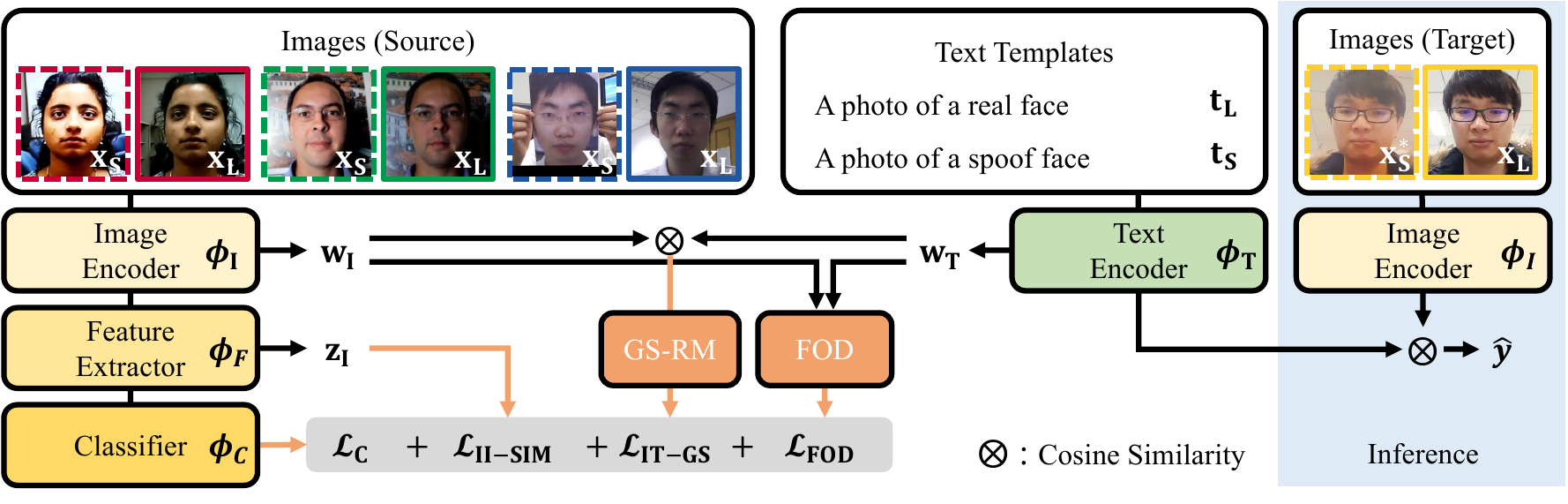}
\vspace{-5mm}
\caption{{\bf{Framework overview}}: Our framework, based on CLIP~\cite{Srivatsan_2023_ICCV}, achieves domain generality by decomposing representations into invariant and specific components within the Text-Image embedding space. 
(1) GS-RM balances group-wise loss across domains to enhance domain invariance. (2) FOD leverages the Gram-Schmidt process to decompose domain-invariant and domain-specific components, effectively mitigating major spurious correlations.
(3) The feature extractor and classifier enhance the embedding of images within the space, ensuring robust feature representation. }
\label{fig:Framework}
\vspace{-3mm}
\end{figure*}
Face Anti-Spoofing (FAS) research aims to enhance the security of biometric authentication in facial recognition systems. Initially, FAS was studied using single-domain datasets~\cite{liu2018learning, yu2020multi, yu2020searching}, but this approach struggles with generalizability when encountering unseen domains. This limitation led to the development of cross-domain FAS approaches,
categorized as follows: Domain Adaptation FAS (DAFAS)~\cite{wang2020cross, zhou2022generative, yue2023cyclically} and Domain Generalizable FAS (DGFAS)~\cite{liu2021dual, liu2021adaptive, wang2022domain, wang2022patchnet, sun2023rethinking, Srivatsan_2023_ICCV}.
DAFAS requires access to target datasets, whereas DGFAS does not. Therefore, DGFAS is more suitable for real-world applications.
\subsection{Domain Generalizable Face Anti-Spoofing.}
DGFAS is to learn common features for domain-agnostic use by training only on labeled source domain datasets.
One study proposed adversarial loss and a triplet loss-based approach~\cite{jia2020single}.
The patch-based method~\cite{wang2022patchnet} makes two different augmented patches from the same image similar for learning good representation. The CLIP-based method~\cite{Srivatsan_2023_ICCV} aligns an ensemble of the class description with image features for improving generality. On the other hand, the instance-level approach~\cite{zhou2023instance} focuses on extracting general representations by reducing the sensitivity of features to varying styles. 

With the growing prominence of Large Language Models (LLMs), recent research has utilized text embedding features derived from LLMs to guide and enhance the extraction of domain-invariant features~\cite{Srivatsan_2023_ICCV, wang2025tf, liu2024cfpl, liubottom}.
Since these methods focus only on domain-invariant features, they struggle to account for domain-specific components embedded within invariant features, such as spurious correlations.

\subsection{Invariant Learning} 
Machine learning models learn complex networks by minimizing training loss through Empirical Risk Minimization (ERM)~\cite{NIPS1991_ff4d5fbb}. However, ERM has a clear limitation for generalization, as models tend to learn biases from training data. To counter this, Invariant Risk Minimization (IRM)~\cite{arjovsky2019invariant, ahuja2021invariance} was proposed, encouraging networks to avoid dataset biases by enforcing an optimal classifier over the representation space that is consistent across multiple environments. Despite its promise, IRM still remains challenging due to the complexity of the optimization problem~\cite{kamath2021does, rosenfeld2021risks}. To handle this, EQRM~\cite{EQRM} proposed a new probabilistic aspect of the optimization problem, assuming target and test domains are drawn underlying (meta) distribution over multiple domains. Invariant learning has been adapted to diverse applications such as face anti-spoofing~\cite{shao2019multi}, semantic segmentation~\cite{gong2019dlow}, and person Re-ID~\cite{jin2020feature,wang2020domainmix}. 

Recently, SAFAS~\cite{sun2023rethinking} proposed training domain-specific classifiers and aligning their directions, demonstrating that Invariant Risk Minimization (IRM) enhances domain invariance in DGFAS. However, SAFAS still suffers from unavoidable domain gaps caused by misaligned domain-specific components, particularly the classifier’s bias terms. In contrast, our approach explicitly decomposes feature representations into domain-invariant and domain-specific subspaces, effectively aligning both weights and biases to mitigate these domain gaps.
\section{Methodology}
\subsection{Problem Setup}
The goal of DGFAS is to detect spoofing or presentation attacks on datasets except those used for training. We denote the image space as $\mathcal{X}=\mathbb{R}^{H\times W\times 3}$. Here, $H$ and $W$ represent the height and width of the image, respectively, with three channels corresponding to the RGB color model to encode color information for each pixel. Each image input is labeled as $\mathcal{Y} = \{\textit{Live}(0),~\textit{Spoof}(1)\}$, and the dataset's domain type is denoted as $\mathcal{E} = \{1, \cdots, e\}$, with a total of $e$ domains. We further derive the dataset $\mathcal{D}=\{(\mathbf{x}, y, e)|\mathbf{x}\in\mathcal{X},y\in\mathcal{Y},e\in\mathcal{E}\}$.
We assume domain $\mathcal{D}$ is given as training data. Then, our goal is to train model $\Phi$:
\begin{equation}
    \Phi:\mathbf{x}^*\rightarrow\{\textit{Live}~(0),~\textit{Spoof}~(1)\},
\end{equation}
where $\Phi$ consists of feature extractor $\phi_{f}$ and classifier $\phi_{c}$. 
$\Phi$ is trained with training data and distinguishes well whether new samples $\mathbf{x}^*$ are lives or not. 
New samples $\mathbf{x}^*$ are defined on an unseen domain $\mathcal{E}^*$.

Fig.~\ref{fig:Framework} illustrates the proposed framework, comprising the text embedding network $\phi_{\mathbf{T}}(\mathbf{t}):\mathbf{x}\rightarrow\mathbf{w}_{\mathbf{T}}$, the image embedding network $\phi_{\mathbf{I}}(\mathbf{x}):\mathbf{x}\rightarrow\mathbf{w}_{\mathbf{I}}$, the feature extractor $\phi_{\mathbf{F}}(\mathbf{w}_{\mathbf{I}}):\mathbf{w}_{\mathbf{I}}\rightarrow\mathbf{z}$, and the classifier $\phi_{\mathbf{C}}(\mathbf{z}):\mathbf{z}\rightarrow\hat{y}$, with Group-wise Scaling Risk Minimization (GS-RM) and Feature Orthogonal Decomposition (FOD) modules. 
During the inference phase, the framework predicts spoofness by calculating the cosine similarity between the text-image feature $\mathbf{w}_\mathbf{T}$ and the image features $\mathbf{w}_\mathbf{I}$ embedded by the text and image encoders. We explain the details regarding the learning losses of the proposed framework in the following sections.
\begin{figure*}[t]
\centering
        \includegraphics[width=1\linewidth]{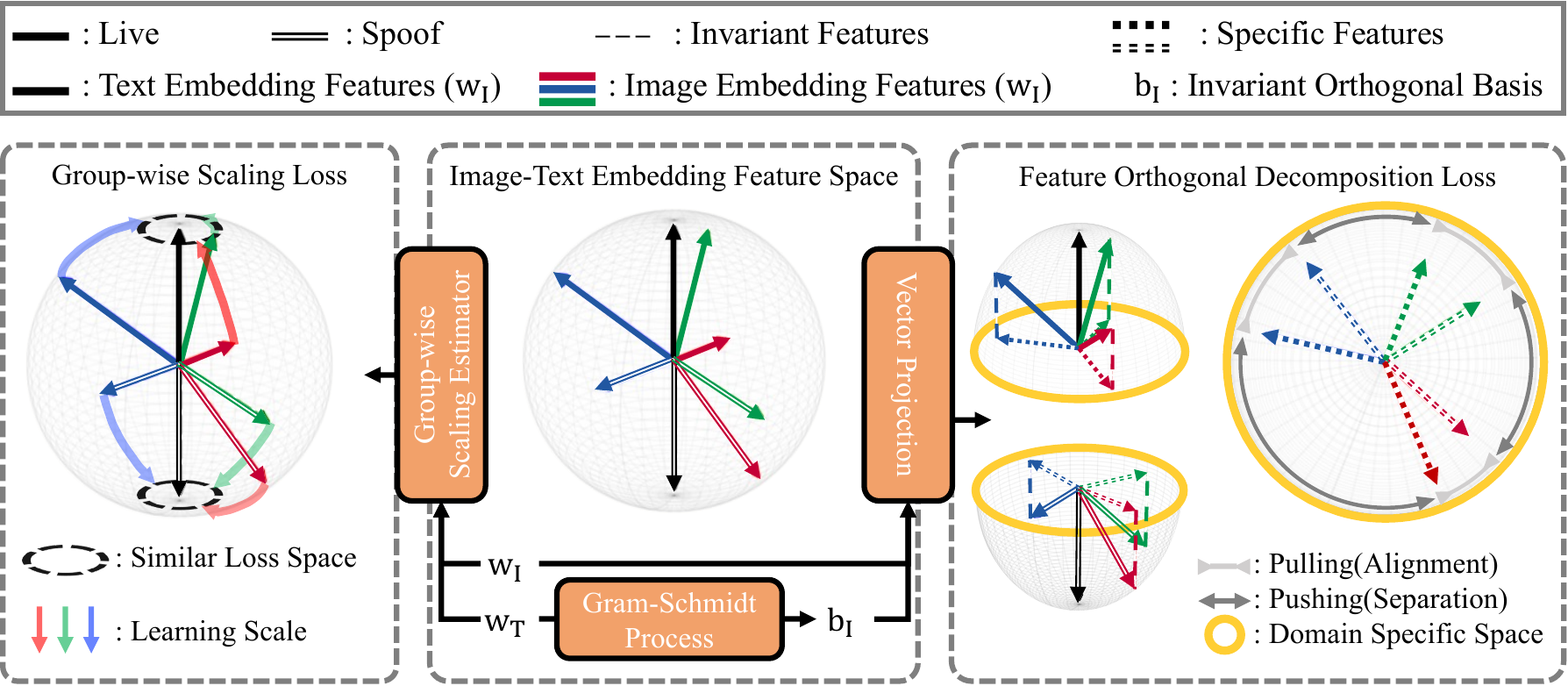}
\caption{{\bf{Insight for GS-RM and FOD}}: The center panel illustrates text features and the features of each domain represented within a hypersphere. The left panel demonstrates GS, which penalizes low-loss (high-similarity) groups and boosts high-loss (low-similarity) groups to achieve balanced learning quantities across groups. The right panel utilizes orthogonal bases and orthogonal projections to decompose domain-specific components from image embedding vectors. Subsequently, FOD aligns the domain-specific components orthogonally to the invariant components by leveraging domain labels during training.}
\vspace{-3mm}
\label{fig:loss}
\end{figure*}
\subsection{Preliminary}

ERM optimizes the network without accounting for domain environments, which can result in the network learning more from certain domains and less from others, finally leading to biased information. Invariant Risk Minimization (IRM)~\cite{arjovsky2019invariant} is proposed to solve the problem of Empirical Risk Minimization. IRM aims to optimize networks across all domains, and the objective formula is defined as follows:
\begin{equation}\label{eq:IRM}
    \begin{aligned}
        \mathcal{L}_{\text{IRM}}:=&\min_{\Phi} \frac{1}{|\mathcal{E}|}\sum_{e\in\mathcal{E}}\mathcal{R}^{e}(\Phi)+{\scriptstyle{||\nabla_{ w|w=1}\mathcal{R}^{e}(w\cdot\Phi)||^{2}}}, \\
        &s.t.~~\forall e \in \mathcal{E},~\Phi\in\arg\min_{\Phi*}\mathcal{R}^{e}(\Phi*),
    \end{aligned}
\end{equation}
where $\mathcal{R}^{e}(\Phi):=\mathbb{E}_{(\mathbf{x},y)\sim\mathcal{D}_{e}}[\ell_{\text{ce}}(\Phi(\mathbf{x}),y)]$ and $\mathcal{D}_{e}:=\{(\mathbf{x},y)|\mathbf{x}\in\mathcal{X},y\in\mathcal{Y},e=e\}$. Eq.~\ref{eq:IRM} forces the network to learn domain invariant features. However, IRM has difficulty in solving the optimization problem.~\cite{rosenfeld2021risks, kamath2021does}. 

To deal with the issue, SAFAS proposed a Projected Gradient (PG-IRM)~\cite{sun2023rethinking}, as follows:

\begin{equation}
    \begin{aligned}
        \mathcal{L}_{\text{PG-IRM}}&:=\min_{\phi_{f},\phi_{c(1)},\cdots,\phi_{c(e)}} \frac{1}{|\mathcal{E}|}\sum_{e\in\mathcal{E}}\mathcal{R}^{e}(\phi_{f},\phi_{c}), \\
        s.t.~~&\exists\phi_{c(e)}\in\arg \min_{\phi_{c(e)}}\mathcal{R}^{e}(\phi_{f},\phi_{c(e)}),\\
        &\forall e \in \mathcal{E},~\phi_{c(e)}\in\gamma_{\alpha}(\phi_{c(e)}),
    \end{aligned}
\end{equation}
where $\gamma_{\alpha}$ means $\alpha$-adjacency set, which is the space where the classifier is Euclidean projected.
PG-IRM trains classifiers $\phi_{c(e)}$ corresponding to each domain $e$ and makes each classifier similar by projecting interpolation. 
Specifically, the classifier aligns the direction of translation between spoofness and liveness. 
However, each classifier becomes strongly entangled with domain-specific components due to overfitting on individual domain datasets, which manifests as a bias term. Consequently, these bias terms result in gaps that degrade generalization performance across domains.

\subsection{Geometrical Motivation\label{sec:discussion}}
Our geometrical motivation aims to improve generality by orthogonally decomposing the feature space into domain-invariant and domain-specific feature spaces, defined as $\mathbf{b_I}$ and $\mathbf{b_S}$, respectively. 
Generality is established as follows: We assume that a feature $\mathbf{f}$ is always combination of both domain-invariant $\mathbf{f_I}$ and domain-specific $\mathbf{f_S}$ components, i.e., $\mathbf{f}=\mathbf{f_I}+\mathbf{f_S}$.
When applying a linear transformation (classifier) computed as $\mathbf{f}\cdot{w}^T$, domain bias arises from $\mathbf{f_S}\cdot{w}^T$  and is compensated by adding a bias term $b$. However, since $\mathbf{f_S}$ induces various local boundaries due to differences across domains, as shown in \cref{fig:teaser}, aligning bias across all domains is challenging. To ensure universal applicability across all domains, two conditions must be satisfied: 
First, $\mathbf{f_I}$ must exist within the orthogonal basis space $\mathbf{b_I}$ derived from optimal $w^{*}$. 
Second, $\mathbf{f_S}$ must exist in $\mathbf{b_S}$ orthogonal to $\mathbf{b_I}$, called domain specific feature space. Under these conditions, $\mathbf{f_S}\cdot{w}^T$ stays zero due to the orthogonality of the inner product, ensuring that $b$ is consistently aligned as zero across domains.

\subsection{Group-wise Scaling Risk Minimization}
We introduce Group-wise Scaling Risk Minimization (GS-RM) to enhance domain-invariance, drawing inspiration from~\cite{pmlr-v202-jung23a}. 
The GS-RM mechanism is designed to achieve alignment in the bias term by ensuring consistent classifier performance across all domains, thereby mitigating overfitting to any specific domain dataset. In contrast, PG-IRM~\cite{Srivatsan_2023_ICCV} enforces alignment among classifier weights that are individually overfitted to their respective domains, consequently leading to misalignment of bias terms across domains.

To induce GS-RM, we first define group-wise loss for each group dataset.
We define group dataset as $\mathcal{D}_{g}=\{(\mathbf{x},y)|\mathbf{x}\in\mathcal{X},(y,e)=g\}$, where $g\in\mathcal{G}=\mathcal{Y}\times\mathcal{E}$.
Then, we formulate group-wise losses as follows:
\begin{equation}
    \mathcal{L}_{g}(\ell_{\text{ce}},\Phi) = \mathbb{E}_{(\mathbf{x},y)\sim\mathcal{D}_g}[\ell_{\text{ce}}(f(\mathbf{x};\Phi),y)],\label{eq:Gw-loss}
\end{equation}
where $\ell_{\text{ce}}$ represents the cross-entropy loss, and $f(\cdot;\cdot)$ denotes the network, of which the input is to the left of the semicolon and the network module to the right.

Second, we normalize group-wise loss to establish clear distinctions in the current iteration. Normalized loss is calculated by subtracting the mean and dividing it by the standard deviation for group-wise losses, defined as follows:

\begin{equation}
    \bar{\mathcal{L}}_{g} = \frac{\mathcal{L}_{g} - \mathbb{E}_{g}[\mathcal{L}_{g}]}{\sqrt{\mathbb{E}_{g}[\mathcal{L}^2_{g}] - \mathbb{E}^2_{g}[\mathcal{L}_{g}]}}\label{eq:norm-Gw-loss}
\end{equation}

Then, we introduce a loss-adaptive scaler, called the Group-wise Scaling Estimator (\cref{fig:loss}), which assigns a higher scaling factor to high-loss scores and a lower scaling factor to low-loss scores.

\begin{equation}
    \sigma(\bar{\mathcal{L}}_{g};\alpha,\beta):=\frac{\beta}{1+exp(-\bar{\mathcal{L}}_{g}/\alpha)}-\frac{\beta}{2}+1,\label{eq:estimator}
\end{equation}
where $\alpha$ and $\beta$ are $\log(|g|)$/2 and 1.5. Each parameter was empirically determined through experimentation.

Finally, we use the group-wise loss (Eq.~\ref{eq:Gw-loss}), group-wise normalized loss (Eq.~\ref{eq:norm-Gw-loss}) and GS Estimator (Eq.~\ref{eq:estimator}) to define GS-RM as follows:
\begin{equation}
    \begin{aligned}
        &\mathcal{L}_{\text{GS-RM}}:=\min_{\Phi} \frac{1}{|\mathcal{G}|}\sum_{g\in\mathcal{G}}\sigma(\bar{\mathcal{L}}_{g}(\Phi))\mathcal{L}_{g}(\Phi) \\
    \end{aligned}    
\end{equation}

\paragraph{Image-Text Similarity Loss}
We define image embedding vector $\mathbf{w}_{\mathbf{I}}:=f(\mathbf{x};\phi_{\mathbf{I}})$ and text embedding vector $\mathbf{w}_{\mathbf{T}}:=f(\mathbf{t};\phi_{\mathbf{T}})$.
Then, we formulate the group-wise image-text similarity loss, defined as follows:
\begin{equation}
    \mathcal{L}_{\text{IT}_{g}} = \mathbb{E}_{(\mathbf{x},y)\sim\mathcal{D}_{g}}[\ell_{\text{ce}}(\mathbf{w}_{\mathbf{I}}\otimes\mathbf{w}_{\mathbf{T}},y)],
\end{equation}
where $\otimes$ means cosine similarity. We further apply GS-RM for the loss to achieve consistent separation of real and spoofed samples across multiple domains. Then, the image-text similarity loss with GS-RM is defined as follows:
\begin{equation}
    \mathcal{L}_{\text{IT-GS}} = \frac{1}{|\mathcal{G}|}\sum_{g\in\mathcal{G}}\sigma(\bar{\mathcal{L}}_{\text{IT}_{g}})\mathcal{L}_{\text{IT}_{g}}
    \label{eq:GS},
\end{equation}
where $\sigma$ denotes GS Estimator (Eq.~\ref{eq:estimator}).

The text embedding vector serves as effective guidance for extracting domain-invariant features~\cite{Srivatsan_2023_ICCV, wang2025tf, liu2024cfpl, liubottom}. Therefore, we utilize this vector as classifier weights, ensuring that it satisfies the first condition mentioned in \cref{sec:discussion} by balancing performance across domains.

\begin{table*}[t!]
\centering
\resizebox{1\linewidth}{!}{
\begin{tabular}{@{}l|c|cc|cc|cc|cc|cc}
\specialrule{.1em}{.05em}{.05em}
\multirow{2}{*}{Method} & \multirow{2}{*}{conference} & \multicolumn{2}{c|}{OCI~$\rightarrow$~M} & \multicolumn{2}{c|}{OMI$\rightarrow$C} & \multicolumn{2}{c|}{OCM$\rightarrow$I} & \multicolumn{2}{c|}{ICM$\rightarrow$O} & \multicolumn{2}{c}{Average} \\ 
& & HTER & AUC & HTER & AUC & HTER & AUC& HTER & AUC &HTER & AUC \\ 
\specialrule{.1em}{.05em}{.05em}
MMD-AAE~\cite{mmd-aae} & CVPR 18 & 27.08 & 83.19 & 44.59 & 58.29 & 31.58 & 75.18 & 40.98 & 63.08 & 36.06 & 69.94 \\
MADDG~\cite{shao2019multi} & CVPR 19 & 17.69 & 88.06 & 24.50 & 84.51 & 22.19 & 84.99 & 27.98 & 80.02 & 23.09 & 84.40 \\
SDA~\cite{sda} & AAAI 21 & 15.40 & 91.80 & 24.50 & 84.40 & 15.60 & 90.10 & 23.10 & 84.30 & 19.65 & 87.65 \\
SSDG-R~\cite{jia2020single} & CVPR 20 & 7.38 & 97.17 & 10.44 & 95.94 & 11.71 & 96.59 & 15.61 & 91.54 & 11.29 & 95.29 \\
SSAN-R~\cite{wang2022domain} & CVPR 22 & 6.67 & 98.75 & 10.00 & 96.67 & 8.88 & 96.79 & 13.72 & 93.63 & 9.82 & 96.46 \\
PatchNet~\cite{wang2022patchnet} & CVPR 22 & 7.10 & 98.46 & 11.33 & 94.58 & 13.40 & 95.67 & 11.82 & 95.07 & 10.90 & 95.95 \\
SAFAS~\cite{sun2023rethinking}   & CVPR 23 & 5.95 & 96.55 & 8.78 & 95.37 & 6.58 & 97.54 & 10.00 & 96.23 & 3.60 & 99.00 \\
FLIP-MCL$^{*}$~\cite{Srivatsan_2023_ICCV} & ICCV 23 & 7.50 & 98.19 & 1.30 & 99.19 & 5.33 & 97.92 & 5.19 & 98.48 & 4.83 & 98.45 \\
CFPL~\cite{liu2024cfpl} & CVPR 24 & 3.09 & 99.45 & 2.56 & 99.10 & 5.43 & 98.41 & 3.33 & 99.05 & 3.60 & 99.00 \\
BUDoPT~\cite{liubottom} & ECCV 24 & \underline{0.95} & \underline{99.70} & 2.85 & 98.03 & 4.40 & 98.54 & \bf{2.26} & 98.78 & 2.62 & 98.76 \\
TF-FAS~\cite{wang2025tf} & ECCV 24 & 3.44 & 99.42 & \bf{0.81} & \underline{99.92} & \bf{2.24} & \bf{99.67} & \bf{2.26} & \bf{99.48} & \underline{2.19} & \bf{99.96} \\
\rowcolor[HTML]{ECF4FF} GD-FAS & Ours & \bf{0.42} & \bf{99.88} & \underline{0.93} & \bf{99.99} & \underline{3.33} & \underline{98.61} & 2.64 & \underline{99.34} & \bf{1.83} & \underline{99.46} \\
\specialrule{.1em}{.05em}{.05em}
FLIP-MCL$^{\dagger}$~\cite{Srivatsan_2023_ICCV} & ICCV 23 & 4.95 & 98.11 & 0.54 & 99.98 & 4.25 & 99.07 & 2.31 & \underline{99.63} & 3.01 & 99.19\\
CFPL$^{\dagger}$~\cite{liu2024cfpl} & CVPR 24 & 1.43 & 99.28 & 2.56 & 99.10 & 5.43 & 98.41 & 2.50 & 99.42 & 2.98 & 99.05 \\
BUDoPT$^{\dagger}$~\cite{liubottom} & ECCV 24 & \underline{0.40} & \underline{99.99} & \underline{0.26} & \underline{99.96} & \bf{1.38} & 99.69 & \underline{1.60} & 99.51 & \bf{0.91} & 99.78 \\
TF-FAS$^{\dagger}$~\cite{wang2025tf} & ECCV 24 & 1.49 & 99.80 & 0.58 & \bf{99.99 }& \underline{1.56} & \bf{99.89} & \bf{1.43} & \bf{99.93} & 1.27 & \bf{99.90} \\
\rowcolor[HTML]{ECF4FF} GD-FAS$^{\dagger}$ & Ours & \bf{0.00} & \bf{100.00}& \bf{0.19} & 99.95 & 1.67 & \underline{99.71} & 1.92 & 99.50 & \underline{0.95} & \underline{99.79} \\
\specialrule{.1em}{.05em}{.05em}
\end{tabular}}
\caption{{\bf{Comparison results on the benchmark}}: Best results considered across epochs. The bold font and the underline represent the best and the second, respectively\label{tab:best}. $\dagger$ indicates the use of an extra source dataset (CelebA-Spoof~\cite{Diwen2014}]), and $*$ means reproducing.}
\end{table*}

\subsection{Feature Orthogonal Decomposition}
SAFAS~\cite{Srivatsan_2023_ICCV} and FOD both aim to align the weight terms across different domains. However, while SAFAS misaligns the bias term across domains, FOD effectively preserves it.
Before describing Feature Orthogonal Decomposition (FOD), we introduce the vector projection to decompose representation. The vector projection is the orthogonal projection of $\mathbf{v}$ onto the line spanned by $\mathbf{u}$ and is defined as:
\begin{equation}\label{eq:proj}
    \text{proj}_{\mathbf{u}}(\mathbf{v})=\frac{<\mathbf{v},\mathbf{u}>}{<\mathbf{u},\mathbf{u}>}\mathbf{u},
\end{equation}
where $<\mathbf{v},\mathbf{u}>$ denotes the inner product of $\mathbf{v}$ and $\mathbf{u}$. The Gram-Schmidt process~\cite{cheney2009linear} is a way of finding a set of two or more vectors that are perpendicular to each other using Eq.~\ref{eq:proj}. It helps to obtain an orthogonal basis $\mathbf{b_{I}}$ from text embedding vectors, treating them as invariant vectors.
We now formulate invariant features as follows:
\begin{equation}
    \mathbf{f_{I}}=\text{proj}_{\mathbf{b_{I}}}(\mathbf{w_{I}})
\end{equation}
Then, we derive the domain-specific feature $\mathbf{f_{S}}$ by removing the domain-invariant feature $\mathbf{f_{I}}$ from the image embedding features $\mathbf{W_{I}}$. As a result, the domain-invariant $\mathbf{f_{I}}$ and domain-specific features $\mathbf{f_{S}}$ become fully orthogonalized, ensuring their inner product is always zero.

Next, we train domain-specific components using these domain-specific features, explicitly aligning the classifier's bias terms across different domains.
\paragraph{Feature Orthogonal Decomposition Loss}
We define the FOD loss within the domain-specific space using domain labels, as follows:
\begin{equation}\label{eq:DAS}
    \mathcal{L}_{\text{FOD}}=\sum_{i=1}^{N}\frac{-1}{|P(i)|}\sum_{p\in P(i)}\log\frac{\exp(\mathbf{f_{S}}^{i}\cdot\mathbf{f_{S}}^{p}/\tau)}{\sum_{j,j\neq i}^{N}\exp(\mathbf{f_{S}}^{i}\cdot\mathbf{f_{S}}^{j}/\tau)},
\end{equation}
where $P(i)$ denotes the positive sample set defined as $P(i):=\{\mathbf{f_{S}}^{j}|\exists e^{j}=e^{i}\}$, $\tau$ represents temperature parameter, and $N$ and superscript indicate the number of batch samples and the index of batch samples, respectively.

\paragraph{Image-Image Similarity Loss and Class Loss}
With the contrastive approach proposed by~\cite{wang2022patchnet} on image feature vector space $\mathcal{Z}\owns \mathbf{z}$, We calculate the InfoNCE loss by contrasting positive pairs derived from the same image with negative pairs sampled from other images.
\begin{equation}
    \mathcal{L}_{\text{II-SIM}} = \sum_{i=1}^{2N}\frac{-1}{|P(i)|}\sum_{j\in P(i)}\log\frac{\exp(\mathbf{z}_{i}\cdot\mathbf{z}_{j}/\tau)}{\sum_{k=1,k\neq i}^{2N}\exp(\mathbf{z}_{i}\cdot\mathbf{z}_{k}/\tau)},
\label{eq:II}
\end{equation}
where $\mathbf{z}$ means feature vector defined as $\mathbf{z}=f(\mathbf{x};\phi_{\mathbf{I}},\phi_{\mathbf{F}})$ and $P(i)$ denotes the positive sample.
Then, we calculate class loss 
\begin{equation}
    \mathcal{L}_{\mathbf{C}}=\mathbb{E}[\ell_{\text{ce}}(f(\mathbf{x};\phi_{\mathbf{I}},\phi_{\mathbf{F}},\phi_{\mathbf{C}}),y)]\label{eq:class}
\end{equation}

\paragraph{Total Loss}Finally, for training the proposed framework, the total loss is as follows:
\begin{equation}
    \mathcal{L}_{\text{Total}} = \mathcal{L}_{\text{IT-GS}}+\lambda_{1}\mathcal{L}_{\text{FOD}} + \lambda_{2}\mathcal{L}_{\text{II-SIM}}+\mathcal{L}_{\text{C}},
\end{equation}
where $\lambda_{1}$ and $\lambda_{2}$ are 0.8 and 0.1 as hyper-parameters, respectively.
Our framework is trained in an end-to-end manner by minimizing the overall objective function.
\section{Experiments}
\subsection{Experimental Setting}
\paragraph{Dataset}
We conduct experiments on four widely used FAS datasets: CASIA~\cite{zhang2012face} (C), OULU-NPU~\cite{OULU_NPU_2017} (O), Idiap Replay-Attack~\cite{Chingovska_BIOSIG-2012} (I), and MSU-MFSD~\cite{Diwen2014} (M). Furthermore, we validate our model on the Spoof in Wild (SiW) Mv2 dataset~\cite{guo2022multi} (W), which represents an in-the-wild setting.
Following the protocol in previous works and SAFAS ~\cite{sun2023rethinking}, we treat each domain as a separate domain and use leave-one-out test protocol to evaluate cross-domain generalization. Specifically in Tab.~\ref{tab:best}, OCI~$\rightarrow$~M implies that the model is trained on O, C and I, while it is tested on M. OMI~$\rightarrow$~C, OCM~$\rightarrow$~I and ICM~$\rightarrow$~O are defined in a same manner.

\begin{figure*}[t!]
\centering
\includegraphics[width=\linewidth]{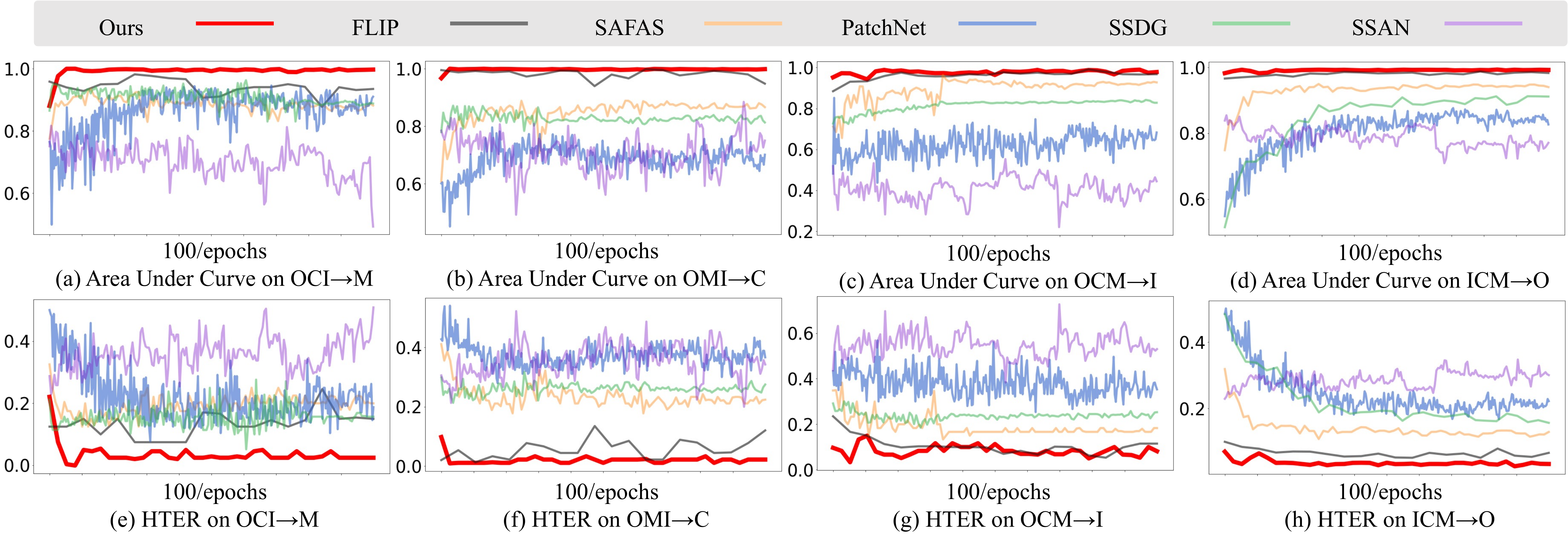}
\caption{{\bf{Learning stability}}: Changes in performance measurements according to epoch settings for each domain.\label{fig:plot}}
\end{figure*}

\paragraph{Implement Details}
The experiments were conducted on the Nvidia RTX A6000. We use 224x224 image size, 16 batch size, Adam optimizer with learning rate 3-e06 and weight decay 1-e06, and StepLR scheduler with step size 10 and gamma 0.1.
To crop the face region used for learning and inference, we adopted the face information via a Multi-Task Convolutional Neural Network (MTCNN)~\cite{zhang2016joint} with a selection of padding value 0.6.

\paragraph{Metrics}
For the evaluation, we use traditional FAS metrics~\cite{sun2023rethinking,Srivatsan_2023_ICCV}, called Halt Total Error Rate (HTER) and Area Under the Curve (AUC). In addition, we introduce another metric for model reliability, called Expected Calibration Error (ECE)~\cite{ECE}.
To measure the reliability, calibration error is defined as the difference between the model's prediction probability and actual probability. Since calculating the actual probability is not feasible, it is approximated using the accuracy of samples, defined as follows:
\begin{equation}
    \mathbf{ECE} = \sum_{m=1}^{M}\frac{|B_{m}|}{n}|\mathbf{acc}(B_{m})-\mathbf{conf}(B_{m})|,
\end{equation}
where $B_{m}$ is $m$-th data bin divided into $M$ equal intervals about predictive probabilities, $\mathbf{acc}(B_{m})$ means the accuracy of $B_{m}$, and $\mathbf{conf}(B_{m})$ means the expectation of prediction probability of $B_{m}$. A lower ECE score indicates greater confidence in the model's predictions.

\begin{table}[t!]
\centering
\resizebox{1\linewidth}{!}{
\begin{tabular}{c|c|c|c|c}
\specialrule{.1em}{.05em}{.05em} 
Method & OCI$\rightarrow$M & OMI$\rightarrow$C & OCM$\rightarrow$I & ICM$\rightarrow$O\\
\specialrule{.1em}{.05em}{.05em}
SSDG-R~\cite{jia2020single} & 8.40 & 11.09 & 8.94 & 9.47 \\
SSAN-R~\cite{wang2022domain} & 9.58 & 5.92 & 12.03 & 11.10 \\
PatchNet~\cite{wang2022patchnet} & 19.98 & 20.98 & 24.31 & 34.91 \\
SAFAS~\cite{sun2023rethinking} & 18.12 & 17.77 & 7.77 & 13.85 \\
FLIP-MCL~\cite{Srivatsan_2023_ICCV} & 3.94 & \textbf{1.45} & 2.62 & 5.45\\
\rowcolor[HTML]{ECF4FF}GD-FAS(Ours) & \textbf{3.39} & 1.82 & \textbf{2.37} & \textbf{2.86}\\
\specialrule{.1em}{.05em}{.05em} 
\end{tabular}}
\caption{{\bf{Model reliability}}: Comparative measurement of Expected Calibration Error (ECE) in various scenarios.\label{tab:ece}}
\end{table}

\subsection{Comparison Results on the Benchmark}
In this section, we provide quantitative comparison results across various approaches, covering different domain shifts to comprehensively evaluate the effectiveness of our methods in diverse scenarios. Following a traditional protocol, we select results based on two criteria: the highest AUC-HTER and the lowest HTER, determined by the corresponding threshold. Regarding the HTER metric, Tab.~\ref{tab:best} demonstrates that our approach achieves state-of-the-art performance in scenarios like OCI→M while attaining the second-best performance in OMI→C and OCM→I. Overall, our method outperforms the second-best method by 0.36 points. When incorporating an extra source dataset, our approach achieves state-of-the-art performance in cases OCI→M and OMI→C. Notably, it produces perfect predictions in the OCI→M scenario.

\subsection{Evaluate Reliability and Bias Term Alignment}
The confidence score is directly influenced by misaligned bias terms across different domains, as they affect the classification threshold. Therefore, a lower Expected Calibration Error (ECE) score indicates better alignment, whereas a higher ECE score reflects poorer alignment. As shown in Table 2, our method outperforms other methods in most cases. Notably, SAFAS exhibits a higher (worse) ECE score, primarily due to misalignment in the classifier’s bias.
Additional calibration results are provided in Appendix~\ref{ap:reliability}.

\begin{table*}[t!]
\centering
\resizebox{1\linewidth}{!}{
\begin{tabular}{cc|ccc|ccc|ccc|ccc}
\specialrule{.1em}{.05em}{.05em}
\multicolumn{2}{c|}{Module} & \multicolumn{3}{c|}{OCI$\rightarrow$M} & \multicolumn{3}{c|}{OMI$\rightarrow$C} & \multicolumn{3}{c|}{OCM$\rightarrow$I} & \multicolumn{3}{c}{ICM$\rightarrow$O} \\ 
GS-RM & FOD & HTER & AUC & ECE & HTER & AUC & ECE & HTER & AUC & ECE & HTER & AUC & ECE \\ 
\specialrule{.1em}{.05em}{.05em}
           &            & 2.92 & 99.35 & 5.60 & 2.22 & 99.59 & 2.02 & 5.17 & 98.19 & 3.55 & 3.98 & 99.09 & 3.21 \\
\checkmark &            & 2.50 & 99.42 & 4.70 & 2.22 & 99.59 & 2.02 & 4.83 & 98.08 & 1.58 & 3.75 & 99.14 & 2.61 \\
           & \checkmark & 2.08 & 99.83 & \bf{1.60} & \bf{0.93} & 99.95 & \bf{1.00} & \bf{3.33} & \bf{99.17} & \bf{1.48} & 3.38 & 99.15 & \bf{2.38}  \\
\checkmark & \checkmark & \bf{0.42} & \bf{99.88} & 3.39 & \bf{0.93} & \bf{99.99} & 1.82 & \bf{3.33} & 98.61 & 2.37 & \bf{2.64} & \bf{99.34} & 2.86 \\
\specialrule{.1em}{.05em}{.05em} 
\end{tabular}}
\caption{{\bf{Ablation study on each component for GD-FAS}}: Quantitative results.\label{tab:ablation}}
\end{table*}

\subsection{Learning Stability}
The convergence stability observed in our model, as shown in Fig.~\ref{fig:plot}, underscores its ability to steadily improve and maintain performance throughout the training epochs. This stability demonstrates the robustness of our approach, thereby providing a reliable solution for real-world applications where domain shifts are frequent.

In contrast, other methods, such as SAFAS, PatchNet, SSDG, and SSAN, exhibit considerable fluctuations during training, indicating their instability in adapting to domain shifts.  Specifically, FLIP, the most recent approach, achieves competitive performance only with early stopping applied, as its performance degrades in later epochs. This is particularly evident in Fig.~\ref{fig:plot}, where FLIP’s HTER performance worsens significantly without early stopping.

Overall, our model's superior stability and consistent convergence, as shown in both AUC and HTER, reflect its effectiveness in handling domain shifts and maintaining reliability over extended training epochs. This highlights the practical utility of our approach in the real world.

\begin{figure}[t!]
\centering
\includegraphics[width=\linewidth]{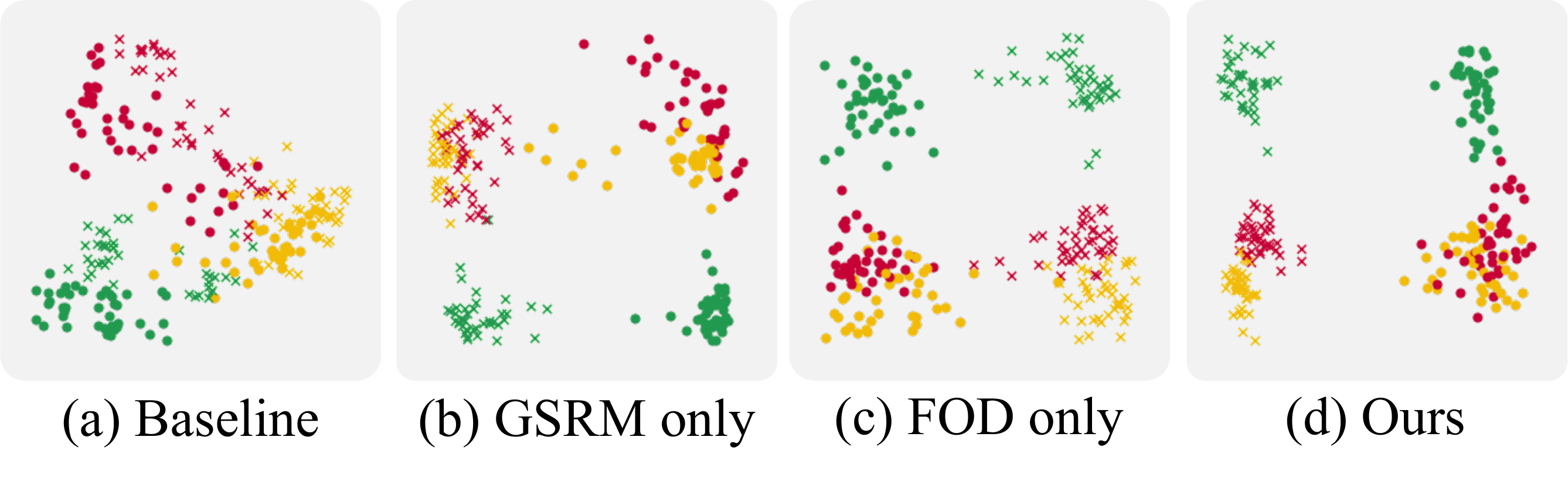}
\caption{{\bf{Ablation study on each component for GD-FAS}}: 2D PCA Visualization for OCI→M according to each module\label{fig:ablation}}
\vspace{-5mm}
\end{figure}

\subsection{Ablation Study}
We conducted an ablation study to evaluate the effectiveness of GSRM and FOD on domain generalization and model reliability, as summarized in Tab.~\ref{tab:ablation}. The results demonstrate that the combination of GSRM and FOD consistently achieves state-of-the-art performance in most scenarios. Notably, the full model achieves the best results in the OCI→M setting (HTER: 0.42\%, AUC: 99.88\%) and the OMI→C setting (HTER: 0.93\%, AUC: 99.99\%). Concerning ECE, there exists a trade-off between model reliability and other performance metrics. While FOD enhances both reliability and generalization, GS-RM primarily focuses on improving generalization. As a result, using only FOD can significantly reduce ECE, but our full model achieves a more balanced trade-off between FAS performance and calibration reliability.

In Fig.~\ref{fig:ablation}, panels (a) and (b) illustrate that GSRM achieves consistent separation of real and spoofed samples across multiple domains, while panels (c) and (d) demonstrate that FOD effectively suppresses domain-specific information within the domain-separable space.

\begin{figure}[t!]
\centering
\includegraphics[width=\linewidth]{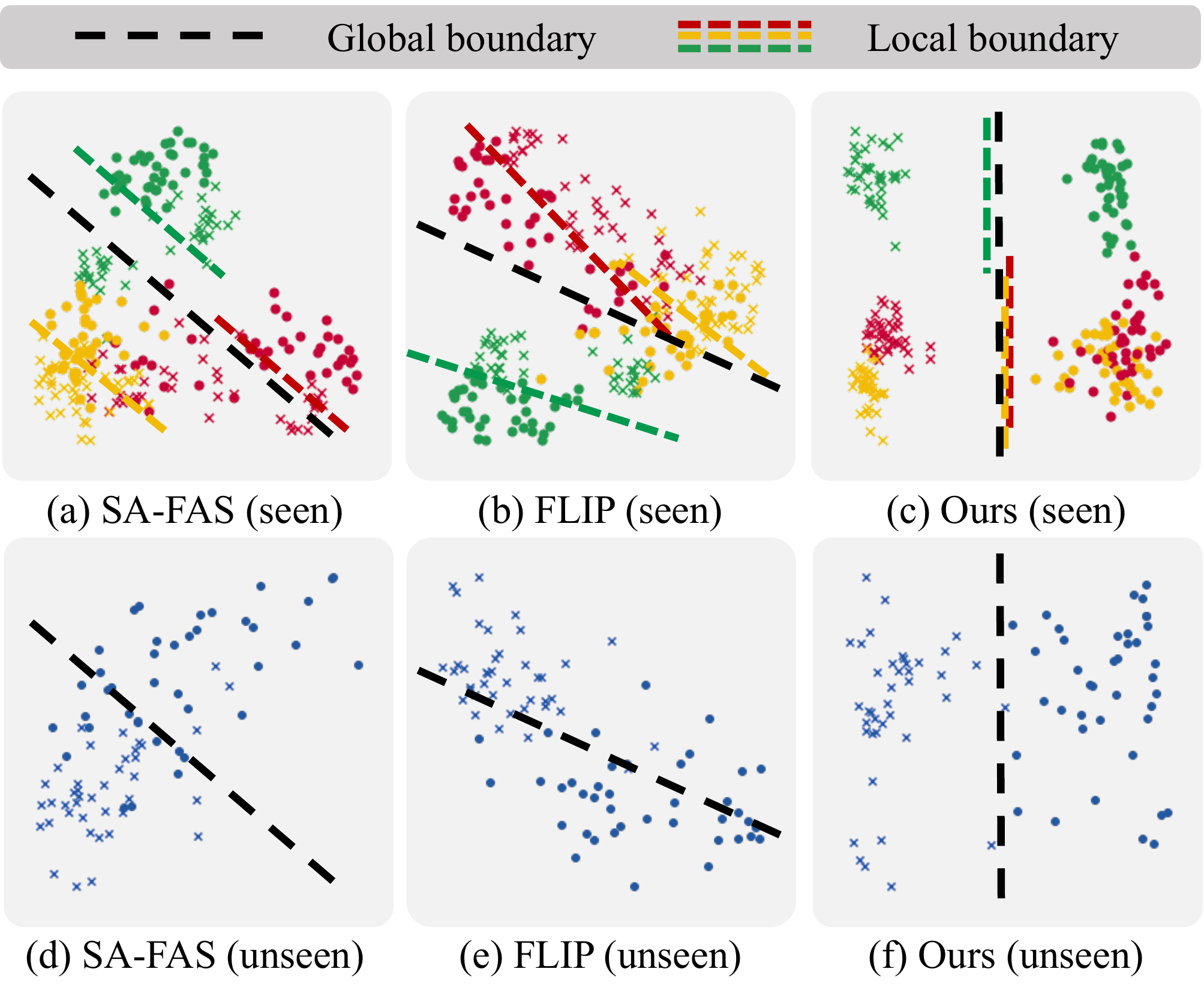}
\caption{{\bf{Comparative Visual Analysis}}: 2D PCA Visualization for OCI→M according to different models \label{fig:visualization}}
\vspace{-5mm}
\end{figure}

\subsection{Visual Analysis}
We employed PCA to preserve the global structure of the data, capturing its overall patterns and directionality for visual analysis. As depicted in Fig.~\ref{fig:visualization}-(c), our method effectively isolates domain-specific information into distinct, separable spaces while maintaining learning consistency across domains. Consequently, as shown in Fig.~\ref{fig:visualization}-(f), our approach establishes a universal boundary that robustly separates previously unseen datasets. In contrast, Fig.~\ref{fig:visualization}-(d) and (e) illustrate that previous models have difficulty in establishing a universal boundary in the unseen domain, resulting in poor separation between spoof and liveness in feature space.
Additional analyses are provided in Appendix~\ref{ap:visual}.

\section{Conclusion}
In this paper, we introduce GD-FAS, a novel Domain Generalizable Face Anti-Spoofing (DGFAS) framework designed to align classifier biases and weights across different domains through Group-wise Scaling Risk Minimization (GS-RM) and Feature Orthogonal Decomposition (FOD).
GS-RM facilitates bias alignment by employing a loss-adaptive scaling factor to balance group-wise losses across multiple domains, ensuring consistent separation of real and spoof samples. FOD leverages the Gram-Schmidt orthogonalization process to explicitly decompose feature representations into domain-invariant and domain-specific subspaces. By enforcing orthogonality between these feature spaces using domain labels, FOD ensures effective alignment of weights across domains without negatively impacting bias alignment.
Extensive experimental evaluations on standard benchmarks demonstrate that GD-FAS achieves state-of-the-art performance, exhibiting superior reliability, reduced bias misalignment, and enhanced generalization stability and adaptability in unseen domain scenarios.

\section*{Acknowledgement}
This work was partly supported by the Institute of Information \& Communications Technology Planning \& Evaluation (IITP) grant funded by the Korea government (MSIT) [IITP-2023(2024)-RS-2024-00418847, Graduate School of Metaverse Convergence support program; RS-2021-II211341, Artificial Intelligence Graduate School Program (Chung-Ang University)].
{
    \small
    \bibliographystyle{ieeenat_fullname}
    \bibliography{main}
}
\clearpage
\maketitlesupplementary
\appendix
\setcounter{figure}{0}
\setcounter{table}{0}
\renewcommand\thefigure{\Alph{figure}}
\renewcommand\thetable{\Alph{table}}

\begin{figure*}[t!]
    \centering
    \includegraphics[width=\linewidth]{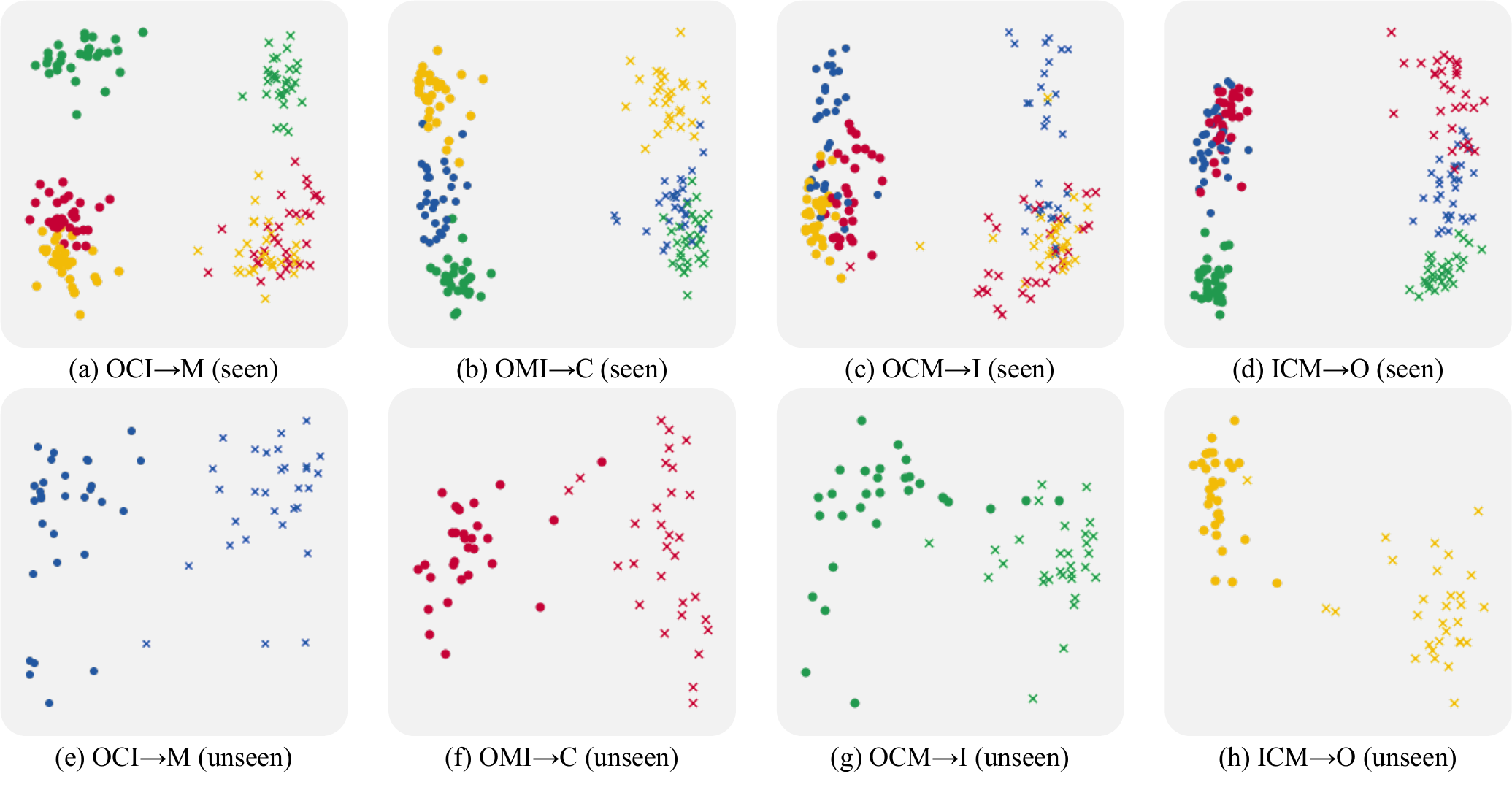}
    \captionof{figure}{{\bf{Visual analysis}}: 2D PCA Visualization across domains. The term (seen) refers to the training dataset, while (unseen) refers to the target dataset that was not encountered during training. {\large$\circ$} and {\large$\times$} represent liveness and spoofness, respectively. The datasets are represented as follows: {\color[HTML]{C50135}{{\large$\circ$}}},{\color[HTML]{C50135}{{\large$\times$}}}: CASIA, {\color[HTML]{21994E}{{\large$\circ$}}},{\color[HTML]{21994E}{{\large$\times$}}}: Idaip Replay-Attack, {\color[HTML]{2257A1}{{\large$\circ$}}},{\color[HTML]{2257A1}{{\large$\times$}}}: MSU-MFSD, {\color[HTML]{F2BB07}{{\large$\circ$}}},{\color[HTML]{F2BB07}{{\large$\times$}}}: OULU-NPU.\label{fig:A_das}}
\end{figure*}

\section{Additional Visual Analysis\label{ap:visual}}
In this section, we describe additional visual analyses that were not covered in detail in Section 4.6 of the main paper. Specifically, we provide experimental validation of our method using PCA-based visualizations.

\begin{figure*}[t!]
    \centering
    \includegraphics[width=\linewidth]{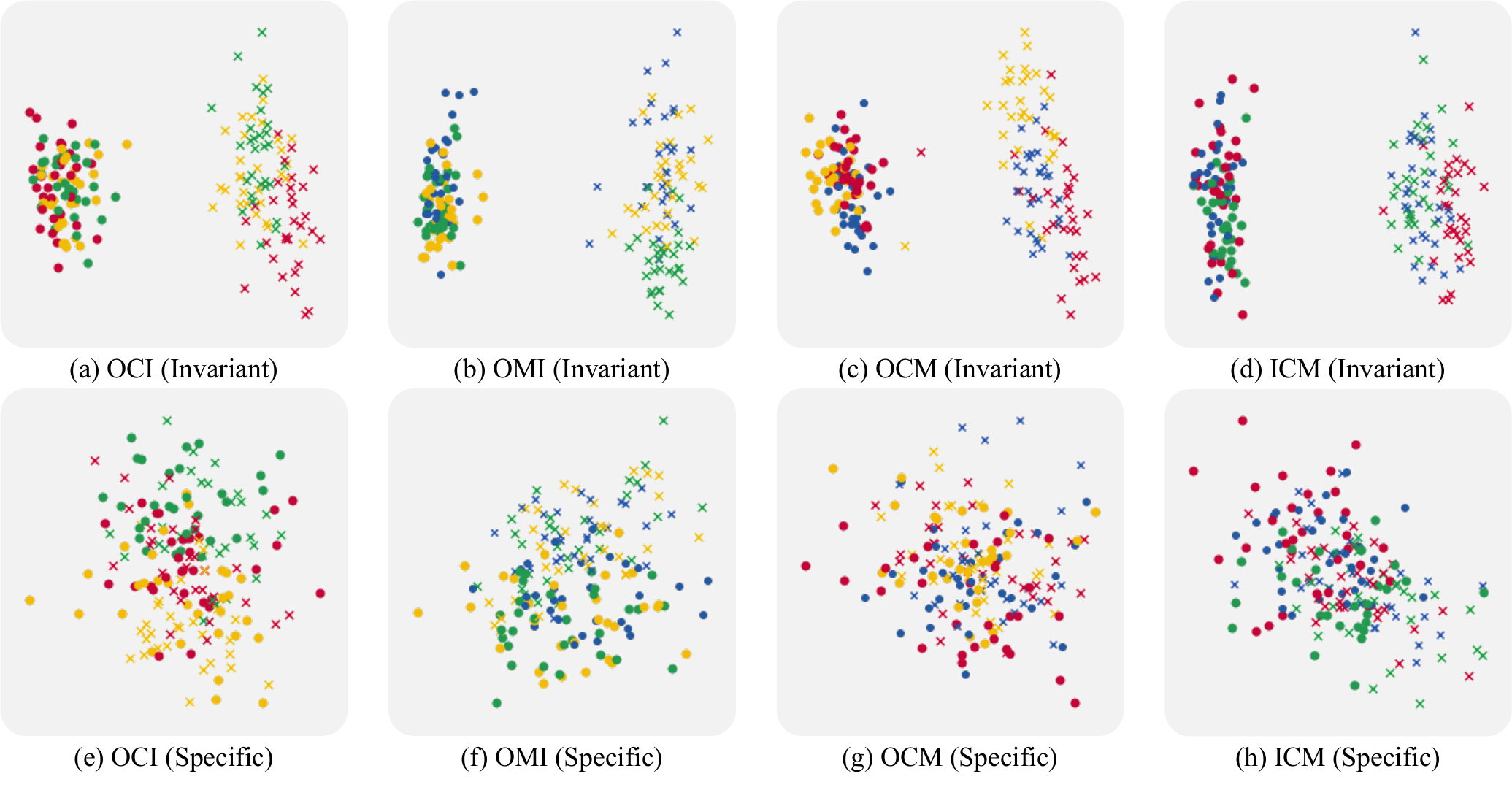}
    \caption{{\bf{Decomposition into invariance and specificity for domain}}: 2D PCA Visualization across domains. The term (Invariant) refers to features projected onto the invariant space, while (specific) refers to features projected onto the domain-specific separable space. {\large$\circ$} and {\large$\times$} represent liveness and spoofness, respectively. The datasets are represented as follows: 
 {\color[HTML]{C50135}{{\large$\circ$}}},{\color[HTML]{C50135}{{\large$\times$}}}: CASIA, {\color[HTML]{21994E}{{\large$\circ$}}},{\color[HTML]{21994E}{{\large$\times$}}}: Idaip Replay-Attack, {\color[HTML]{2257A1}{{\large$\circ$}}},{\color[HTML]{2257A1}{{\large$\times$}}}: MSU-MFSD, {\color[HTML]{F2BB07}{{\large$\circ$}}},{\color[HTML]{F2BB07}{{\large$\times$}}}: OULU-NPU.\label{fig:A_is}}
\end{figure*}

\begin{figure*}[t!]
    \centering
    \includegraphics[width=\linewidth]{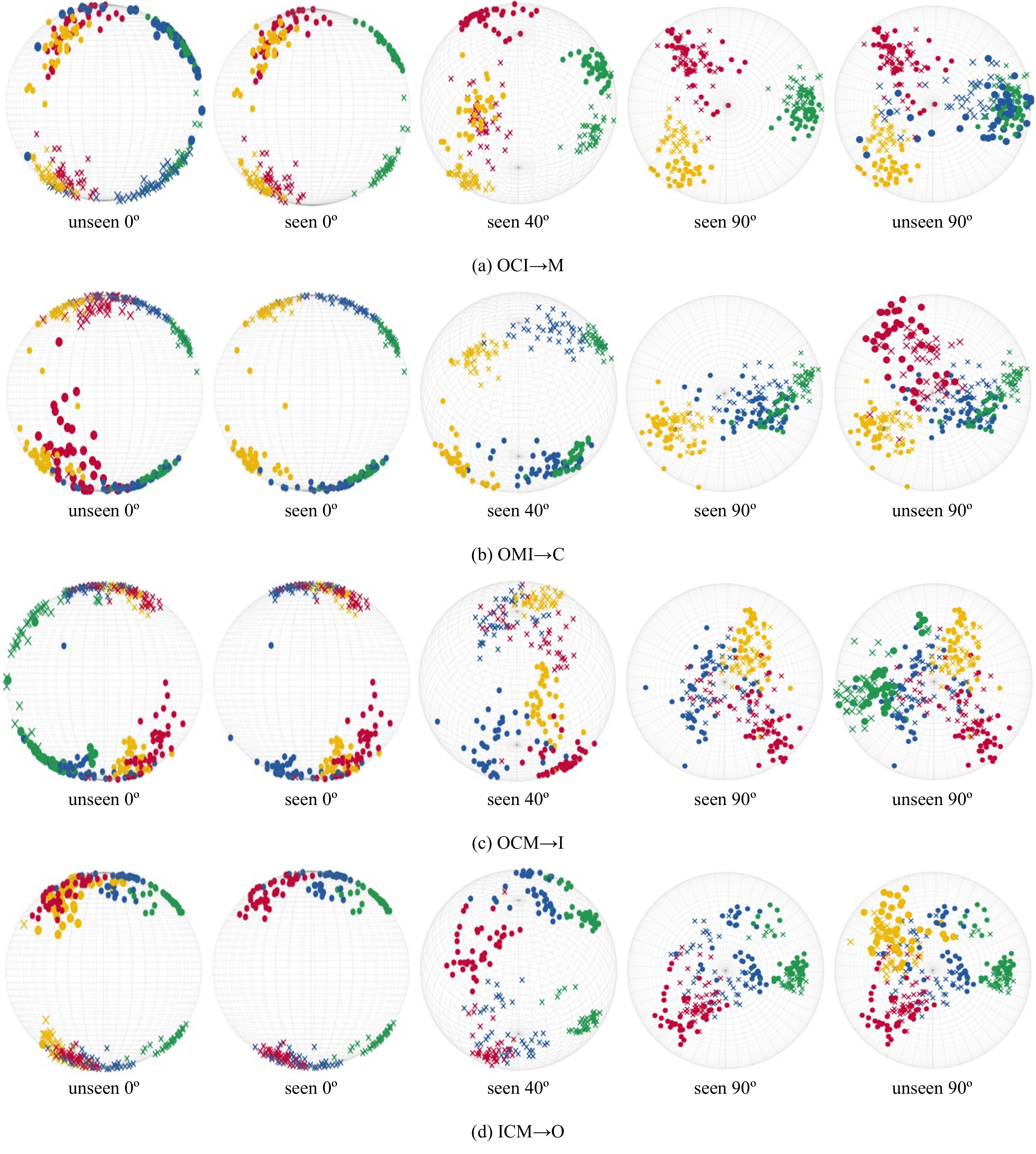}
    \captionof{figure}{{\bf{3D visualization projected on sphere}}: PCA 3D Visualization across domains. The terms (0º), (40º), and (90º) represent the angles of rotation about the vertical axis.  {\large$\circ$} and {\large$\times$} represent liveness and spoofness, respectively. The datasets are represented as follows: {\color[HTML]{C50135}{{\large$\circ$}}},{\color[HTML]{C50135}{{\large$\times$}}}: CASIA, {\color[HTML]{21994E}{{\large$\circ$}}},{\color[HTML]{21994E}{{\large$\times$}}}: Idaip Replay-Attack, {\color[HTML]{2257A1}{{\large$\circ$}}},{\color[HTML]{2257A1}{{\large$\times$}}}: MSU-MFSD, {\color[HTML]{F2BB07}{{\large$\circ$}}},{\color[HTML]{F2BB07}{{\large$\times$}}}: OULU-NPU.\label{fig:A_3d}}
    \vspace{5mm}
\end{figure*}

\subsection{Feature Orthogonal Decomposition\label{sec:A_DAS}}
For PCA analysis and visualization, we calculate the 2D principal components based on the source dataset (training data) which indicates the (seen) domains. The target dataset (test data) which indicates the (unseen) domains is then projected onto these 2D principal components. Finally, we visualize the features of the source and target datasets projected onto the derived principal components.

As shown in Fig.~\ref{fig:A_das}-(a), (b), (c), and (d), all cases demonstrate strong learning consistency. However, there are some mismatches between the domain-separable feature space and the decision boundary, indicating that they are not parallel to each other. Specifically, Fig.~\ref{fig:A_das}-(b) exhibits little skewed decision boundary in the Idaip domain, (c) in the CASIA domain, and (d) in the MSU domain.
Referring to Tab 1 in the main paper, the alignment quality correlates with the HTER scores. The most aligned case, (a), achieves the lowest HTER of 0.42, followed by (b) with 0.93, (d) with 2.64, and finally, (c), which has the highest HTER of 3.33.

Moreover, Fig.~\ref{fig:A_das}-(e), (f), (g), and (h) further confirm that better alignment quality enhances the alignment of the domain-specific separable space in the target dataset.

This analysis validates that aligning the domain-specific feature space with the decision boundary significantly contributes to improving domain generality.

\subsection{Decomposition into Invariant and Specific Features\label{sec:A_IS}}
For PCA analysis and visualization, we conduct the 2D principal components analysis for invariant features which are projected onto the invariant basis features. The invariant basis features are driven by Gram-Schmidt process~\cite{cheney2009linear}. Specific features are then obtained by subtracting the invariant features from the original features. Finally, we visualize the invariant features and specific features separately to analyze their behavior.

The domain-invariant features should demonstrate that they are challenging to distinguish across different domains. In Fig.~\ref{fig:A_is}-(a), (b), (c), and (d), this is evident as the liveness features are difficult to associate with specific domains. In contrast, spoofness features show some level of domain separability. This might be from the fact that liveness typically represents a single characteristic, while spoofness encompasses a variety of attack types, such as print attacks, video attacks, and partial attacks. However, they still prove their invariance to domains.

Specific features, on the other hand, should neither clearly distinguish spoofness from liveness nor differentiate between domains, as they are unrelated to domain invariance. Moreover, since they are projected onto the principal components of domain-invariant features, domain separability should also be minimal. As expected, Fig.~\ref{fig:A_is}-(e), (f), (g), and (h) depict indistinct patterns, where neither spoofness nor domain information can be identified.

The overall visualizations confirm that our method effectively and explicitly decomposes domain-invariant components from domain-specific components, achieving the intended decomposition.

\subsection{3D visualization}
We aim to intuitively examine the domain-invariant and domain-specific components. To achieve this, we further conduct 3D principal components analysis and project the features onto them for visualization, as shown in Fig.~\ref{fig:A_3d}. In the figure, the x-axis refers to the horizontal axis when the sphere is viewed from the front. Accordingly, at 0º on the x-axis, the domain-invariant components are highlighted; at 90º, the domain-specific components are emphasized; and at 40º, both components are visible. Additionally, the seen visualization represents data from the training set, while the unseen visualization includes both target and training datasets.

As proposed in Sec.~\ref{sec:A_DAS}, Fig.~\ref{fig:A_3d} (seen, 0º) demonstrates that learning consistency is well-aligned. Furthermore, as discussed in Sec~\ref{sec:A_IS}, Fig.~\ref{fig:A_3d} (seen, 90º) shows that the domain-specific components can distinguish domains but cannot differentiate between liveness and spoofness. These results confirm that the domain-invariant and domain-specific components are effectively and explicitly separated. We also provide 3D visualized GIF images, including ours, FLIP, and SAFAS.

\begin{figure*}[t!]
    \centering
    \includegraphics[width=0.98\linewidth]{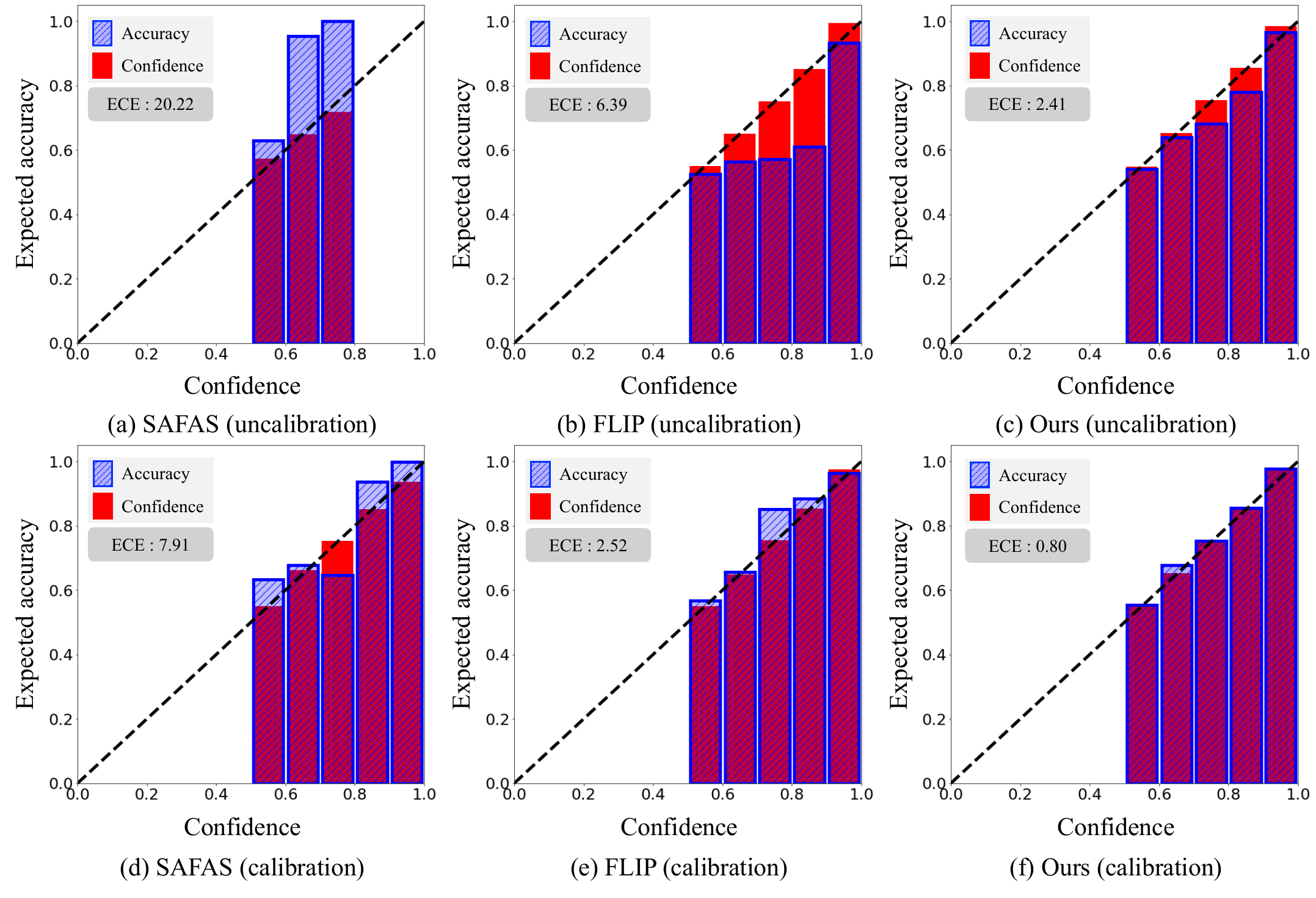}
    \caption{{\bf{Reliable Diagram}}: Comparison with Previous Methods. The terms (uncalibration) and (calibration) are distinguished based on whether the uncertainty has been calibrated.\label{fig:A_ece}}
\end{figure*}

\begin{table*}[t!]
\centering
\resizebox{1\linewidth}{!}{
\begin{tabular}{c|ccc|ccc|ccc|ccc}
\specialrule{.1em}{.05em}{.05em} 
Methods   & {\small C$\rightarrow$I} & {\small C$\rightarrow$M} & {\small C$\rightarrow$O} 
          & {\small I$\rightarrow$C} & {\small I$\rightarrow$M} & {\small I$\rightarrow$O} 
          & {\small M$\rightarrow$C} & {\small M$\rightarrow$I} & {\small M$\rightarrow$O}
          & {\small O$\rightarrow$C} & {\small O$\rightarrow$I} & {\small O$\rightarrow$M} \\
\hline
FLIP-MCL$^{\dagger}$~\cite{Srivatsan_2023_ICCV} & 10.57 & 7.15 & 3.91 & 0.68 & 7.22 & 4.22 & 0.19 & 5.88 & 3.95 & 0.19 & 5.69 & 8.40 \\
BUDoPT$^{\dagger}$~\cite{liubottom} & 4.33	 & 2.62	& 4.98 & 0.48 &	1.83 & 4.14	& \textbf{0.00} & 2.45 & 1.87 & 0.44 & 2.53 & 1.43 \\
TF-FAS$^{\dagger}$~\cite{wang2025tf} & 3.06 & \textbf{1.59} & 3.78 & 0.69 & \textbf{1.34} & \textbf{2.50} & 0.11 & 2.31 & \textbf{1.40} & 1.5 & 4.02 & 1.59 \\
\rowcolor[HTML]{ECF4FF} GD-FAS$^{\dagger}$
          & \textbf{2.98} & 3.92 & \textbf{3.36} & \textbf{0.19} & 2.50 & 3.36 & \textbf{0.00} & \textbf{2.23} & 2.59 & \textbf{0.00} & \textbf{1.83} & \textbf{0.42} \\
\specialrule{.1em}{.05em}{.05em}
\end{tabular}}
\caption{\bf{Leave-three-domains-out protocol}: $\dagger$ indicates the use of an extra source dataset (CelebA-Spoof~\cite{Diwen2014}]).\label{tab:l3do}}
\end{table*}

\section{Model Reliability\label{ap:reliability}}
Regardless of how advanced AI models become, the implementation of fundamental safety mechanisms remains essential for their deployment in real-world scenarios, particularly within security systems. This necessity arises because model reliability is often expressed through the probability values associated with predictions; however, recent AI models have demonstrated a tendency toward overconfidence~\cite{guo2017calibration}. Unfortunately, this critical aspect has been largely overlooked in prior studies.

In this section, we aim to underscore the importance of model reliability and delineate how our approach differs from previous methods by discussing the Expected Calibration Error (ECE) and the reliability diagrams introduced in Sec 4.1 of this paper.

A reliability diagram is an analytical tool that visually displays the discrepancy between predicted probability values and the actual probabilities. We conduct a visual analysis of the dataset both with and without uncertainty calibration, utilizing a post-hoc uncertainty calibration method that employs the temperature scaling technique. For this purpose, we split the test dataset into a validation set and a test set with a 2:8 ratio. The validation set is used to calibrate the reliability, while the test set is used to evaluate.

As shown in Fig.~\ref{fig:A_ece}, our method maintains a slightly overconfident state, whereas FLIP is significantly overconfident and, conversely, SAFAS is underconfident. In the calibrated diagrams below, the slightly under and overconfident regions are notably reduced. Here, we observe that our method before calibration outperforms other methods even after calibration.

\begin{table*}[t!]
\centering
\setlength{\tabcolsep}{12pt}
\begin{tabular}{c|c|c|c|c|c|>{\columncolor[HTML]{ECF4FF}}c}
\specialrule{.1em}{.05em}{.05em}
\multicolumn{2}{c|}{Methods} & SSAN-R~\cite{wang2022domain} & SSDG-R~\cite{jia2020single} & DGUA-FAS~\cite{hong2023domain} & BUDoPT~\cite{liubottom} & GD-FAS \\
\hline
\multirow{2}{*}{MI$\rightarrow$C} & HTER & 25.56 & 19.86 & 19.22 & 5.33 & {\textbf{2.22}} \\
                                  & AUC  & 83.89 & 86.46 & 86.81 & 98.92 & \textbf{99.15}\\
\hline
\multirow{2}{*}{MI$\rightarrow$O} & HTER & 24.44 & 27.92 & 20.05 & 5.94 & \textbf{3.75} \\
                                  & AUC  & 82.86 & 78.72 & 88.75 & 98.37 & \textbf{98.75}\\
\specialrule{.1em}{.05em}{.05em}
\end{tabular}
\caption{{\bf{Leave-two-domains-out protocol}}\label{tab:l2do}}
\end{table*}

\begin{table*}[t!]
\centering
\setlength{\tabcolsep}{12.8pt}
\begin{tabular}{c|cc|cc|cc|cc}
\specialrule{.1em}{.05em}{.05em} 
\multirow{2}{*}{Methods} & \multicolumn{2}{c|}{CS$\rightarrow$W} & \multicolumn{2}{c|}{SW$\rightarrow$C} & \multicolumn{2}{c|}{CW$\rightarrow$S} & \multicolumn{2}{c}{Average}\\
&HTER&AUC&HTER&AUC&HTER&AUC&HTER&AUC\\
\specialrule{.1em}{.05em}{.05em}
FLIP-MCL$^{\dagger}$~\cite{Srivatsan_2023_ICCV} & 4.46 & \textbf{99.16} &  9.66 & 96.69 & 11.71 & 95.21 & 8.61 & 97.02 \\
CFPL$^{\dagger}$~\cite{liu2024cfpl} & \textbf{4.40} & 99.11 &  8.13 & 96.70 &  8.50 & 97.00 & 7.01 & 97.60 \\
\rowcolor[HTML]{ECF4FF}GD-FAS$^{\dagger}$ & 7.88 & 97.60 & \textbf{4.01} & \textbf{98.59} & \textbf{5.87} & \textbf{98.04} & \textbf{5.92} & \textbf{98.08} \\
\specialrule{.1em}{.05em}{.05em}
\end{tabular}
\caption{{\bf{Leave-one-domain-out protocol with different datasets}}: $\dagger$ indicates the use of an extra source dataset (CelebA Spoof)\label{tab:l1do}}
\end{table*}

\section{Additional Quantitative Experiments}
We conducted experiments under more challenging settings, including the leave-two-domain-out (Sec.~\ref{sec:l3do}) and leave-three-domain-out (Sec.~\ref{sec:l2do}) protocols. Additionally, we evaluated our method on a different DGFAS dataset to assess its generalizability beyond the original benchmark(Sec.~\ref{sec:CSW}).
We evaluate all experiments using two standard metrics: Half Total Error Rate (HTER) and Area Under the ROC Curve (AUC).

\subsection{Leave-three-domains-out protocol\label{sec:l3do}}
Tab.~\ref{tab:l3do} presents the results under the leave-three-domains-out protocol, where the model is trained on a single domain and evaluated on the remaining unseen domains. Our method, GD-FAS, consistently outperforms existing state-of-the-art approaches across most domain shifts. Notably, it achieves the best performance in particularly challenging scenarios such as O→C and M→C(HTER: 0.00\%) and I→C (HTER: 0.19\%). These results highlight the strong generalizability of our approach across highly diverse and difficult domain pairs.

\subsection{Leave-two-domains-out protocol\label{sec:l2do}}

Tab.~\ref{tab:l2do} presents the performance under the leave-two-domains-out protocol, where the model is trained on two domains and evaluated on the remaining unseen domains—consistent with the experimental setup used in BUDoPT~\cite{liubottom}.
In both MI→C scenarios, GD-FAS achieves the lowest HTER (2.22\% and 3.75\%) and the highest AUC (99.15\% and 98.75\%), significantly outperforming prior methods such as BUDoPT~\cite{liubottom}, DGUA-FAS~\cite{hong2023domain}, and SSDG-R~\cite{jia2020single}. These results demonstrate the effectiveness of our group-wise scaling and orthogonal decomposition techniques in capturing domain-invariant features, even under limited training conditions and complex inter-domain shifts.

\subsection{Different dataset\label{sec:CSW}}
Our experiments were conducted on three diverse datasets—Surf (S)~\cite{zhang2020casia}, CeFA (C)~\cite{ liu2021casia}, and WMCA (W)~\cite{george2019biometric}—under the leave-one-domain-out protocol to evaluate cross-dataset generalization.

As shown in Tab.~\ref{tab:l1do}, while FLIP-MCL and CFPL achieve high AUC scores, GD-FAS demonstrates the most balanced performance, achieving the best average results across all domains (HTER: 5.92\%, AUC: 98.08\%). Notably, in the SW→C setting, GD-FAS outperforms all baselines with a significantly lower HTER (4.01\%) and higher AUC (98.59\%), highlighting its robustness against challenging cross-domain shifts.

\section{Discussion of Computational Cost}
In this section, we discuss the computational cost of GD-FAS.
\paragraph{Training Efficiency} The additional computational overhead introduced by the GS-RM and FOD losses is approximately 10M FLOPs, which is negligible compared to the overall training complexity of FLIP-MCL (88.6G FLOPs). GD-FAS also adopts the same backbone architecture as FLIP-MCL and simply replaces the original MSE loss with the proposed GS-RM and FOD losses. As a result, GD-FAS maintains a comparable training cost to FLIP-MCL. In contrast, prior methods often rely on significantly more resource-intensive components, such as large language models (LLMs) or two-stage training pipelines.
\paragraph{Inference Efficiency} GD-FAS achieves an inference time of 0.01 seconds per frame, matching the computational cost of FLIP-MCL due to their shared inference pipeline. In comparison, other baseline methods typically incur slightly higher computational overhead and latency.

\begin{table*}[t!]
\centering
\setlength{\tabcolsep}{13pt}
\begin{tabular}{c|c|c|c|c}
\specialrule{.1em}{.05em}{.05em} 
 & OCI$\rightarrow$M & OMI$\rightarrow$C & OCM$\rightarrow$I & ICM$\rightarrow$O \\ 
Prediction & HTER~/~AUC~/~ECE & HTER~/~AUC~/~ECE & HTER~/~AUC~/~ECE & HTER~/~AUC~/~ECE\\
\specialrule{.1em}{.05em}{.05em}
\rowcolor[HTML]{ECF4FF} $\mathbf{w_I}\otimes\mathbf{w_T}$& \bf{0.42~/~99.88~/~3.39} & \bf{0.93~/~99.99~/~1.82} & \bf{3.33~/~98.61~/~2.37} & \bf{2.64~/~99.34~/~2.86} \\
$\Phi_{C}$& 2.50~/~99.44~/~5.02 & 2.22~/~99.71~/~5.15 & 4.83~/~98.75~/~3.51 & 3.33~/~99.28~/~2.41 \\     
\specialrule{.1em}{.05em}{.05em}
\end{tabular}
\caption{{\bf{Embedding Features ($\mathbf{w_I}\otimes\mathbf{w_T}$) vs. Classifier ($\phi_C$)}} \label{tab:rub-fc}}
\end{table*}

\begin{table*}[t!]
\centering
\setlength{\tabcolsep}{15pt}
\begin{tabular}{c|c|c|c|c}
\specialrule{.1em}{.05em}{.05em}
batch & OCI$\rightarrow$M & OMI$\rightarrow$C & OCM$\rightarrow$I & ICM$\rightarrow$O \\ 
size & HTER~/~AUC~/~ECE & HTER~/~AUC~/~ECE & HTER~/~AUC~/~ECE & HTER~/~AUC~/~ECE \\ 
\specialrule{.1em}{.05em}{.05em}
3  & 5.00~/~98.77~/~6.38 & 3.15~/~99.24~/~15.81
     & 6.67~/~98.72~/~2.24 & 2.96~/~99.24~/~9.41 \\

8  & 2.50~/~99.85~/~3.60 & 2.22~/~98.84~/~8.67 
     & 5.00~/~98.94~/~2.78 & 3.01~/~99.20~/~5.41 \\  

\rowcolor[HTML]{ECF4FF} 16 & {\bf{0.42}}~/~99.88~/~3.39 & {\bf{0.93}}~/~{\bf{99.99}}~/~1.82 
     & {\bf{3.33}}~/~98.61~/~2.37 & \bf{2.64}~/~\bf{99.34}~/~2.86 \\

24 & {\bf{0.42}}~/~{\bf{99.90}}~/~2.79 & {\bf{0.93}}~/~99.86~/~0.80 
     & {\bf{3.33}}~/~{\bf{99.25}}~/~2.79 & 2.96~/~{\bf{99.34}}~/~5.77\\

32 & {\bf{0.42}}~/~99.73~/~{\bf{2.53}} & 1.11~/~99.76~/~{\bf{0.72}} 
     & 3.50~/~99.24~/~{\bf{1.84}} & 3.01~/~99.07~/~6.72\\
\specialrule{.1em}{.05em}{.05em}
\end{tabular}
\caption{{\bf{Influence of Batch-size}} \label{tab:rub-batch}}
\end{table*}

\begin{table*}[t!]
    \centering
    \setlength{\tabcolsep}{17.5pt}
    \begin{tabular}{c|c|cccc}
    \specialrule{.1em}{.05em}{.05em}
    Method & Target & OCI$\rightarrow$M & OMI$\rightarrow$C & OCM$\rightarrow$I &ICM$\rightarrow$O \\
    \hline
    Gaussian Noise & Image & 7.5 & 3.52 & 3.5 & 4.77\\
    Random Vector & Text & 7.08 & 4.44 & 11.83 & 7.36\\
    a Pohoto of \{dog or cat\} & Text & 3.3 & 12.04 & 5.60 & 15.09\\
    a Photo of \{live of fake\} face & Text & 2.5 & 1.11 & 5.17 & 4.07\\
    \rowcolor[HTML]{ECF4FF}Baseline (FLIP-MCL) & None & \textbf{0.42} & \textbf{0.93} & \textbf{3.33} & \textbf{2.64} \\
    \specialrule{.1em}{.05em}{.05em}
    \end{tabular}
    \caption{\textbf{Robustness across Image and Text Quality}\label{tab:it}}
\end{table*}

\section{Classifier vs Embedding Features}
In this section, we explain why GD-FAS detects spoofing attacks using embedding features rather than relying on the classifier output, as illustrated in Fig.~\ref{fig:Framework}. As shown in Tab.~\ref{tab:rub-fc}, predictions based on embedding features outperform those based on the classifier. This performance gap arises because the embedding space explicitly separates domain-invariant and domain-specific information through orthogonal decomposition. In contrast, the classifier operates on features prior to this decomposition, failing to fully exploit the decomposed representations.

\section{Analysis of Batch-Size Influence}
We analyze the influence of batch size on model performance.
CLIP-based methods typically use a batch size of 3 per domain due to the large model size, resulting in an effective batch size equal to the per-domain batch size multiplied by the number of domains.

As shown in Tab.~\ref{tab:rub-batch}, a batch size of 16 is sufficient for stable and effective training. Increasing the batch size beyond 16 yields marginal gains while requiring GPU memory in excess of 48GB, which may not be practical for most setups.

\begin{figure}[t!]
    \centering
    \includegraphics[width=1\linewidth]{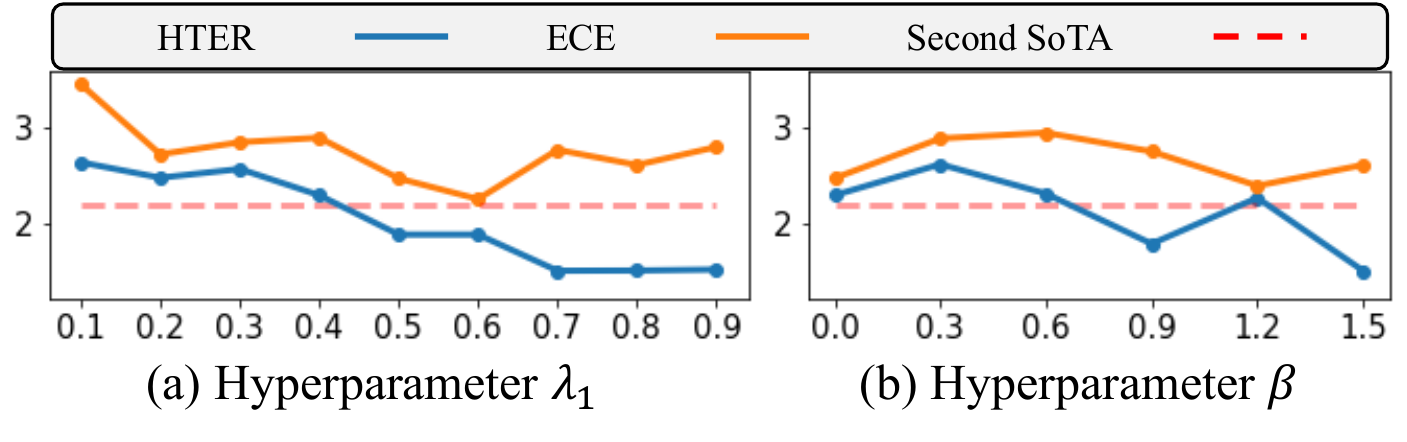}
\caption{\textbf{Hyperparameter Sensitivity}: Default ${\lambda}_1$=0.8, $\beta$=1.5}
\label{fig:hyper}
\end{figure}

\section{Robustness across Image and Text Quality}
We analyze the robustness of GD-FAS to variations in both image and text quality. GD-FAS utilizes the same text templates as FLIP-MCL, comprising six distinct templates for each class—spoof and live—to guide representation learning.
To evaluate robustness, we conduct experiments under four distinct conditions: one related to image quality and three to text quality.
For the image case, we apply Gaussian noise to the input images to simulate visual degradation.
For the text cases:
(1) we replace text embeddings with random vectors,
(2) we use only a single incorrect-class template, and
(3) we use only a single correct-class template.

As descriebed in Tab.~\ref{tab:it}, while both image and text degradations lead to moderate performance drops, degradation in text quality—particularly the use of random vectors or incorrect-class templates—results in a more substantial decline. These findings suggest that text embeddings play a crucial role in enabling GD-FAS to extract domain-invariant features effectively.

\section{Sensitivity of Hyperparameter}
We performed a series of analytical experiments to examine how FOD and GS-RM contribute to overall performance, as shown in Fig.~\ref{fig:hyper}. The hyperparameters $\lambda_1$ and $\beta$ correspond to FOD and GS-RM, respectively.
While our method exhibits slight sensitivity to hyperparameter variations, the combined use of FOD and GS-RM consistently improves performance as their contributions are strengthened. We selected the final hyperparameter values based on HTER, as it directly reflects the model’s domain generalization ability.

\end{document}